\documentclass[dvipsnames]{article}
\usepackage{graphicx}
\usepackage{booktabs}
\usepackage{xcolor}
\usepackage{caption}
\usepackage{subcaption}
\usepackage{showlabels}
\usepackage{enumitem}
\usepackage[ruled,vlined]{algorithm2e}

\usepackage{amsmath,amsfonts,bm}

\def\Eqref#1{Equation~\ref{#1}}
\def\eqref#1{\Eqref{#1}}

\def\1{\bm{1}}
\newcommand{\train}{\mathcal{D}}

\DeclareMathAlphabet{\mathsfit}{\encodingdefault}{\sfdefault}{m}{sl}
\SetMathAlphabet{\mathsfit}{bold}{\encodingdefault}{\sfdefault}{bx}{n}

\DeclareMathOperator*{\argmax}{arg\,max}

\DeclareMathOperator{\Tr}{Tr}

\usepackage{subcaption}
\usepackage{bbm}
\usepackage{mathtools}
\usepackage{amsthm, amssymb}

\def\1{\bm{1}}

\newcommand{\pp}[1]{\left( #1 \right)}

\newcommand{\expect}[2]{\mathbbm{E}_{#1}\left[ #2 \right]}

\PassOptionsToPackage{numbers, compress}{natbib}

\usepackage[preprint]{neurips_2020}

\usepackage[utf8]{inputenc} 
\usepackage[T1]{fontenc}    
\usepackage{hyperref}       
\usepackage{url}            
\usepackage{booktabs}       
\usepackage{amsfonts}       
\usepackage{nicefrac}       
\usepackage{microtype}      
\usepackage{hyperref}

\SetKwFunction{Range}{range}
\SetKwInput{KwInput}{Input}
\SetKwInput{KwParams}{Parameters}

\def\train{{\text{train}}}
\def\val{{\text{val}}}
\def\all{{\text{all}}}
\def\ensemble{{\text{ensemble}}}
\def\eq#1{(\ref{#1})}

\title{Towards NNGP-guided Neural Architecture Search}

\author{%
  Daniel S. Park\thanks{Equal contribution.} \,\,\,\,
  Jaehoon Lee$^*$ \,\,\,\,
  Daiyi Peng\,\,\,\,
  Yuan Cao\,\,\,\,
  Jascha Sohl-Dickstein \\
  \\[2pt]
  Google Research, Brain Team \\[5pt]
  \texttt{\{danielspark, jaehlee, daiyip, yuancao, jaschasd\}@google.com} \\
}

\begin{document}

\maketitle

\begin{abstract}
The predictions of wide Bayesian neural networks are described by a Gaussian process, known as the Neural Network Gaussian Process (NNGP). Analytic forms for NNGP kernels are known for many models, but computing the exact kernel for convolutional architectures is prohibitively expensive. One can obtain effective approximations of these kernels through Monte-Carlo estimation using finite networks at initialization. Monte-Carlo NNGP inference is orders-of-magnitude cheaper in FLOPs compared to gradient descent training when the dataset size is small. Since NNGP inference provides a cheap measure of performance of a network architecture, we investigate its potential as a signal for neural architecture search (NAS). We compute the NNGP performance of approximately 423k networks in the NAS-bench 101 dataset on CIFAR-10 and compare its utility against conventional performance measures obtained by shortened gradient-based training. We carry out a similar analysis on 10k randomly sampled networks in the mobile neural architecture search (MNAS) space for ImageNet. We discover comparative advantages of NNGP-based metrics, and discuss potential applications. In particular, we propose that NNGP performance is an inexpensive signal independent of metrics obtained from training that can either be used for reducing big search spaces, or improving training-based performance measures.
\end{abstract}

\section{Introduction}

The behavior of deep neural networks often becomes analytically tractable when the network width is very large 
\citep{neal, williams1997, hazan2015steps,  schoenholz2016deep, lee2018deep, matthews2018, matthews2018b_arxiv, Borovykh2018, garriga2018deep, novak2018bayesian, yang2017mean, yang2018a, yang2019scaling, yang2019wide, pretorius2019expected, neuraltangents2020, hron2020,Jacot2018ntk, li2018learning, allen2018convergence, du2018gradient, du2018gradienta, zou2018stochastic, Zou2019, chizat2019lazy, lee2019wide, arora2019on, du2019graph, sohl2020infinite, Huang2020OnTN, pennington2017resurrecting,  xiao18a, xiao2019disentangling, Hu2020Provable, li2019enhanced, Arora2020Harnessing, Shankar2020NeuralKW,cho2009kernel, daniely2016toward, poole2016exponential, chen2018rnn, li2018on, daniely2017sgd,  pretorius2018critical, hayou2018selection, karakida2018universal, blumenfeld2019mean, hayou2019meanfield}.
One such example is that Bayesian inference in the parameter space of a deep neural network becomes equivalent to a Gaussian process prediction in function space \citep{neal,lee2018deep,matthews2018}. 
An intriguing consequence of this correspondence is that once the neural network Gaussian process (NNGP) kernel is computed, exact Bayesian inference is possible in wide networks. 
This further allows the computation of properties such as the expected validation accuracy of a wide Bayesian neural network with a given architecture and initialization scale.

We expect performance metrics of the NNGP associated with a given network to correlate well with the network's actual performance after training, since we expect the predictions of Bayesian and gradient descent trained networks to be correlated.
An important topic in neural architecture search (NAS) \cite{zoph2016nas,baker2016designing} is the discovery of computationally cheap methods to predict the fully-trained performance of a given network. 
This suggests NNGP performance should provide a useful signal for NAS. This use case has been previously suggested \citep{novak2019neural, arora2019exact, shankar2020neural}, but never explored.

There have been extensive studies on computing the exact kernel of NNGPs, but the actual computation can be very expensive \cite{novak2019neural, arora2019exact, shankar2020neural}, requiring hundreds of accelerator hours. Furthermore, networks in neural network search spaces typically use operations for which the NNGP correspondence has no known closed form. However, Monte Carlo estimates of NNGP kernels are often far cheaper, and can be computed for any architecture by repeated random initialization of the network \citep{lee2018deep}.

\setcounter{footnote}{0} 
We have released a Colab notebook\footnote{\url{https://github.com/google-research/google-research/tree/master/nngp_nas}} demonstrating our algorithm to compute NNGP performance, as well as metrics used in this paper to measure the quality of NNGP performance as a predictor of the ground-truth network performance.

\subsection{Summary of contributions}

We examine the utility of NNGP validation accuracy for predicting final network performance, and compare it against that of shortened-training, which is a common method for approximating network performance \cite{zoph2016nas}. We do so in two different settings: on the NAS-Bench-101 dataset~\cite{ying2019nasbench101} with 423k network architectures evaluated on CIFAR-10 \cite{cifar10}; and on 10k randomly sampled networks from the MNAS search space \cite{tan2019mnasnet} evaluated on ImageNet \cite{imagenet}. 
In both cases we find that NNGP performance is indicative of final network performance, while being at least an order of magnitude cheaper to compute than shortened-training. 
We further find that for the large NAS-Bench-101 search space, thresholding by NNGP accuracy dramatically shrinks the search space which needs to be examined by more expensive methods. We also demonstrate that NNGP and shortened-training performance can potentially be combined to produce a metric with improved predictive quality on NAS-Bench-101.

\subsection{Further related work}

Proxy tasks, i.e., computationally manageable tasks for approximating the performance of a neural network, are commonly used in neural architecture search \cite{zoph2016nas, tan2019mnasnet, real2019regularized, ghiasi2019fpn}. Early stopping by leveraging features and training curves from finished trials has also been used \cite{swersky2014freezethaw, domhan2015speeding, klein2017learning,baker2017accelerating}. 
A measure for predicting network performance at initialization has recently been proposed in \cite{mellor2020neural}. 
Architecture search can also be framed as finding a sub-network within a super-network that is trained once. In architecture search with weight sharing \cite{pham2018ena, liu2019darts, cai2018proxylessnas}, a controller samples sub-networks from the super-network and uses a per mini-batch reward as an approximation for the sub-network performance. In differentiable architecture search \cite{liu2019darts} the sub-network selection is part of the gradient based training.  Learning-based methods \cite{wen2019NeuralPF} have also been recently applied to predict the performance of the network, where a neural network that takes the architecture as the input and produces the ground truth performance as the output is trained on a subset of the search space.

Our approach is orthogonal and could be combined with these prior works; the NNGP performance studied is obtained without any gradient based training and furthermore, without any pre-existing gradient based training data.

\section{Background}

\subsection{NNGP Inference}

Consider a deep neural network with parameters $\theta$, whose architecture maps an input $x$ into a feature vector $\tilde{z} = f(x; \theta)$ of dimension $d$, followed by a linear readout layer with weight variance $\sigma_w^2$ producing predicted labels $y$. Consider $n$ data points $x^1, \cdots, x^n$. In the NNGP approximation, the distribution over output vectors $\tilde{y}_I^1, \cdots, \tilde{y}_I^n$ for any label $I$ at initialization is jointly Gaussian,
\begin{equation}
(\tilde{y}^1_I, \cdots, \tilde{y}^n_I) \sim \mathcal{N}(0, \sigma_w^2 \mathbf{K})\,,
\quad \mathbf{K}_{ij} = K(x^i, x^j) \,
= \expect{\theta}{\frac{1}{d} \sum_{k=1}^d \tilde{z}^i_k \tilde{z}^j_k},
\label{eq K expect}
\end{equation}
where $K(x^i, x^j)$ is the sample-sample second moment of $\tilde{z}$ averaged over random network initializations and units. This approximation has been shown to be exact in the wide limit for many architectures \cite{neal, schoenholz2016deep, lee2018deep, matthews2018, matthews2018b_arxiv,  garriga2018deep, novak2018bayesian, yang2019scaling, yang2019wide, hron2020}. For the rest of the paper, we proceed with the assumption that this is an adequate approximation for the architectures being studied. While exact convergence of Bayesian inference on the neural network parameter space to the NNGP has not been proven for architectures with some of components we use in this work (e.g., max-pooling), it is expected that very general classes of architectures will exhibit GP-like behaviour as they become wide.  

Since the distribution of label vectors are given by a Gaussian distribution, we can compute the exact conditional expectation values of labels. 
If the parameter initialization distribution is interpreted as a prior distribution, this corresponds to the predictions that would be made by a Bayesian neural network. 
In other words, if the inputs $x^1, \cdots, x^{N_\text{train}}$ produce the label vectors $y^1, \cdots, y^{N_\text{train}}$, and introducing a regularization constant $\epsilon$, the expected label vector for an input $x^*$ is
\begin{align}
\mathbb{E}[y^*_I \, |\, (x^i, y^i), x^*] &= 
\sum_{i,j} K(x^*, x^i) \,
(\mathbf{K} + \epsilon \mathbb{I}_{N_\text{train}})^{-1}_{ij}
\, y^j_I\,, \\
\text{Label of $x^*$} &=  \argmax_I \left( \sum_{i,j} K(x^*, x^i) \,
(\mathbf{K} + \epsilon \mathbb{I}_{N_\text{train}})^{-1}_{ij}
\, y^j_I \right) \,.
\end{align}

The central challenge in carrying out an NNGP calculation is computing the kernel $K$. In this paper, we estimate the kernel by a Monte-Carlo method first studied in~\cite{novak2018bayesian}. That is, $K(x^i, x^j)$ is computed by stochastically evaluating the expectation in equation \eq{eq K expect} over repeated random initializations of the network. We denote the number of random initializations the {\it ensemble number}.

We search over a range of normalization constants $\epsilon$, as the result of NNGP inference can vary significantly with $\epsilon$. We normalize the search range of this constant with respect to the average eigenvalue $\lambda$ of the kernel matrix, i.e., $\epsilon = \tilde{\epsilon} \lambda$. Since $\tilde{\epsilon}$ has been made dimensionless in this way, the same search range of $\tilde{\epsilon}$ can be used for all NNGP experiments. We take this range to be \texttt{numpy.logspace(-7, 2, 20)}.

The full procedure for computing NNGP validation accuracy in shown in Algorithm \ref{a:NNGP}. The target label vectors $y^i$ are derived from the target labels $I^i$ to be the one-hot vector shifted to have mean-zero.

\begin{algorithm}[t]
\SetAlgoLined
\KwInput{Training input, label vector pairs $(x^i, y^i)$;
validation input, label pairs $(\bar{x}^a, \bar{I}^a)$;
network function $f(\cdot\,;\,\cdot)$; initialization distribution $p$.}
\KwResult{The NNGP validation accuracy $A_\val$.}
 Initialize $K^{vt} = \mathbf{0}_{N_\val \times N_\train}$,
 $K^{tt} = \mathbf{0}_{N_\text{train} \times N_\text{train}}$,
 $A_\val = 0$\;
 \For{$k \in [0,n_\ensemble)$}{
  Initialize parameters $\theta \sim p\pp{\theta}$\;
  Initialize batch-norm parameters of network $f(\cdot \,;\, \theta))$\;
  \lFor{$i \in [0,N_\train)$}{$z_i = f(x^i; \theta)$}
  \lFor{$a \in [0,N_\val)$}{$\bar{z}_a = f(\bar{x}^a; \theta)$}
  \lFor{$(i,j) \in [0,N_\train)^2$}{$K^{tt}_{ij} \mathrel{+}= {1 \over n_\text{ensemble}d} z_i \cdot z_j$}
  \lFor{$(a,i) \in [0,N_\val) \times [0,N_\train)$}{
  $K^{vt}_{ai} \mathrel{+}= {1 \over n_\text{ensemble}d}\bar{z}_a \cdot z_i$}
  }
 $\lambda = {1 \over N_\train}\Tr(K^{tt})$\;
 \For{$\tilde{\epsilon} \in [\tilde{\epsilon}_1, \cdots, \tilde{\epsilon}_r]$}{
   $Y_{aI} = \sum_{i,j}K^{vt}_{ai}
   (K^{tt} + {\tilde{\epsilon} \lambda} \mathbf{1}_{N_\train})^{-1}_{ij} y^{j}_I$\;
   $C_\val = \sum_a (\bar{I}^a \,\text{==}\, \argmax_I Y_{aI})$\;
   $A_\val = \max\left(A_\val,~{C_\val/ N_\val}\right)$\;
 }
 \caption{NNGP Inference with ensemble number $n_\ensemble$}
 \label{a:NNGP}
\end{algorithm}

\subsubsection*{Computational Cost}
Let $F_A$ be the architecture ($A$) dependent number of flops for inference on a single sample, $d_A$ the dimension of the feature space of the penultimate layer of the network, $n_\ensemble$ the ensemble number, $L$ the number of labels, and $N_\train$ and $N_\val$ the size of the training and validation sets for the NNGP, respectively. The computational FLOPs required for computing the NNGP validation accuracy is
\begin{equation}\label{NNGP flops}
\text{NNGP Flops} = n_\text{ensemble} F_A  N_{\train + \val}
+ {1 \over 3} r N_\train^3
+ 2(d_An_\text{ensemble} + 2Lr ) N_\train N_{\train + \val} \,.
\end{equation}
The details of this computation can be found in section \ref{ap:cc} of the supplementary material (SM). This cost does not scale well with $N_\train$, which we cap at 8k samples when computing NNGP accuracy to keep the inference cost reasonable.

Let us denote the FLOPs required for training a model $A$ per step per sample by $G_A$. We have computed $F_A$ and $G_A$ for all the networks in the NAS-Bench-101 dataset for multiple batch sizes and found that the relation $G_A = 2 F_A$ holds robustly (see section \ref{ap:cc} of the SM). Thus, the cost of training a model for $E$ epochs and carrying out inference for validation can be written as
\begin{equation}\label{training flops}
  \text{Training Flops} = 2 E F_A N^\all_\train + F_A N^\all_\val \,.
\end{equation}
We add the superscript ``all" to emphasize that gradient-based training of the networks is always performed on the entire dataset, while NNGP inference is performed on sub-sampled datasets.

For various plots regarding NAS-Bench-101, we will plot the computational cost of obtaining NNGP or shortened training performance based on average FLOPs computed over all networks in the dataset. We note that $\mathbb{E}_A[F_A] = 2.51$ GFLOPs while $d_A = 512$ for all networks in NAS-Bench-101.

\subsection{Metrics}\label{ss:metrics}

We use the following metrics to evaluate the quality of NNGP and short-training proxy tasks:

\textbf{Kendall Rank Correlation Coefficient (Kendall's Tau): }
The Kendall rank correlation coefficient measures how well the prediction of two orderings agree. Let us assume two orderings of a set. For every pair of elements of the set, let us denote the number of concordant pairs (pairs that are ordered the same way in both orderings) $P$, the number of discordant pairs (pairs that are ordered the opposite way in the two orderings) $Q$, and the number of ties in each of the orderings $T$ and $U$. There are multiple versions of Kendall's Tau which treat ties differently. We use the following definition:
\begin{equation}
\tau = {(P -Q) / \sqrt{(P+Q+U)(P+Q+T)}} \,.
\end{equation}
We compute how well a validation accuracy ranks the networks by computing its Kendall rank correlation coefficient against the ground-truth accuracy of the networks.

\textbf{Correlation Coefficient: }
We compute the Pearson correlation coefficient between the performance metrics and the ground-truth accuracy.

\textbf{Prediction Quality for Exceedance of Threshold Performance  $p_T$ (PQETP-$p_T$): }
To judge the utility of a performance metric of a network, we can measure how well it predicts whether the true network performance is above some threshold $p_T$. We do so by computing the area under the receiver operating characteristics (AUROC). The ROC curve for a binary classifier with a continuous output is obtained by plotting the true positive rate against the false positive rate as the true/false boundary of the classifier is varied. In our case, the binary classifier is set to be the performance metric we would like to evaluate (e.g., validation accuracy after shortened-training, NNGP validation accuracy), and the binary class is whether the ground-truth validation accuracy of the network exceeds the threshold $p_T$. Thus a metric with a better PQETP for performance $p_T$ is better at determining whether the ground-truth performance of the network is above $p_T$.

\textbf{Discovered Performance: }
The ``discovered performance" of a set of networks is obtained by first choosing the top-$k$ performers in the set according to the reference performance metric, and taking the best ground-truth performance among those of the selected $k$ networks, i.e., it is the performance of the top performer ``discovered" using the metric. We choose $k$ to be 10 throughout the paper, and will be computing the average discovered performance of subsets of NAS-Bench-101 of fixed size.

\section{Experiment Design}

\subsection{The NAS-Bench-101 Dataset}

The NAS-Bench-101 dataset \cite{ying2019nasbench101} consists of $\sim$ 423k image classification networks (details of which are provided in section \ref{ap:nasbench} of the supplementary materials) evaluated on CIFAR-10, with the standard train/validation/test split of 40k/10k/10k samples. The dataset contains the validation and training accuracy of each network after training for 4, 12, 36 and 108 epochs for 3 different trials. We take the ground-truth performance of a network to be the average validation accuracy for the three 108-epoch training trials. Our goal is to evaluate how well NNGP performance predicts this ground truth accuracy.

We compute the NNGP validation accuracy with a range of ensemble numbers and fixed sub-sampled training sets and validation sets of different sizes with balanced labels for all networks. We take $N_\train \in \{100, 500, 2000, 8000\}$, $N_\val \in \{ 500, 2000, 5000, 10000 \}$ and $n_\text{ensemble} \in \{ 1, 2, 4, 8, 16, 32\}$. We compare the utility of the NNGP validation accuracy obtained using different values of $(N_\train, N_\val, n_\text{ensemble})$ with those obtained by shortened training. To do so, we use the validation accuracy computed at the end of a single trial of 4, 12 and 36 epoch training. We can compute the cost for obtaining each measure using equations \eq{NNGP flops} and \eq{training flops}. The computational costs will be presented in peta-FLOPs (PFLOPs). All metrics introduced in section \ref{ss:metrics} are computed for the validation accuracies obtained from NNGP inference and shortened training.

\subsection{The MNAS Search Space}

The MNAS search space \cite{tan2019mnasnet} is intended for mobile neural networks on ImageNet \cite{imagenet} with the aim of balancing performance and computational cost. Throughout this paper, we refer to the provided 50k ``validation set" of ImageNet as the ``test set" and split a separate 50k subset from the 1.3M-sample training set, designating it as the validation set for evaluation. We select a randomly sampled set of 10k networks from the MNAS search space to study. The 5-epoch validation accuracy, and the MNAS reward function, which is a function of this accuracy and latency, are computed for all 10k networks. For more details, see section \ref{ap:mnas} of the supplementary material.

We carry out NNGP inference on a random 8k-sample subset of the training set with balanced labels and the 50k validation set. We also construct training and validation subsets using 100 randomly subsampled labels with sizes (1k, 5k) and (8k, 5k). We use these three pairs of training/validation sets for NNGP inference with fixed ensemble number 4, and compare its utility as a performance measure against the MNAS reward obtained by 5-epoch training. We do so by selecting the top-10 networks according to the NNGP validation accuracies and shortened-training measures, and evaluating the ground-truth performance by training the networks for 400-epochs and evaluating the test set accuracy. The hyperparameters used for 5 and 400-epoch training are identical with those used in \cite{tan2019mnasnet}.

\section{Results}

\subsection{CIFAR10 on NAS-Bench-101}

\begin{figure}[h!]
\centering
\includegraphics[width=0.9\columnwidth]{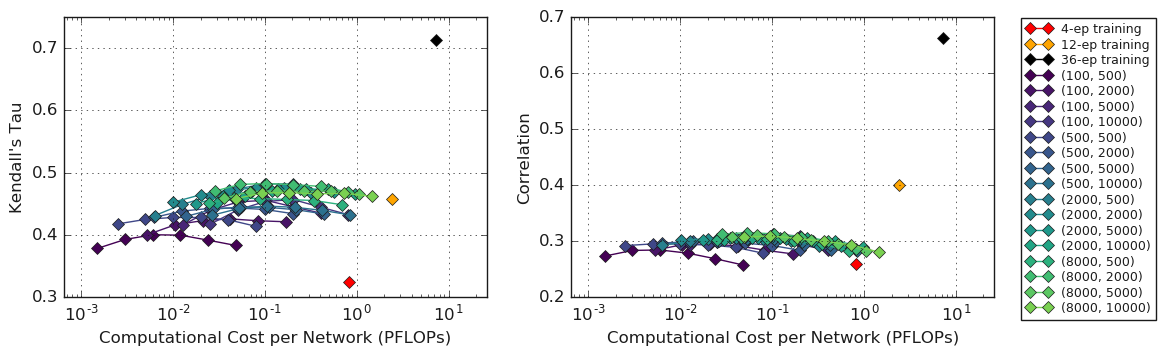}
\caption{
\textbf{NNGP validation accuracy is a computationally cheap way to predict final network performance. } 
Kendall's Tau and Pearson correlation between NNGP/shortened-training validation accuracy and the ground-truth network performance plotted against the computational cost per network for all networks in NAS-Bench-101. Color indicates $(N_\train, N_\val)$ for the NNGP performance, while the different points in each connected plot represents different ensemble numbers. See SM section~\ref{ap:raw} for average prediction performance.}
\label{f:tau R2}
\vskip -0.1in
\end{figure}

We compute Kendall's Tau and the correlation of the NNGP validation accuracies for different combinations of $(N_\train, N_\val, n_\text{ensemble})$ and validation accuracies measured for shortened-training against the ground-truth performance of the $\sim$ 423k networks in NAS-Bench-101. (The raw inference results are presented in SM section~\ref{ap:raw}.) These measures have been plotted against the relative computational flops in figure \ref{f:tau R2}. We find that the Kendall's Tau performance of 4-epoch or 12-epoch training can be matched or bested with NNGP inference with an order-of-magnitude less FLOPs.

\begin{figure}[h!]
\centering
\includegraphics[width=0.98\columnwidth]{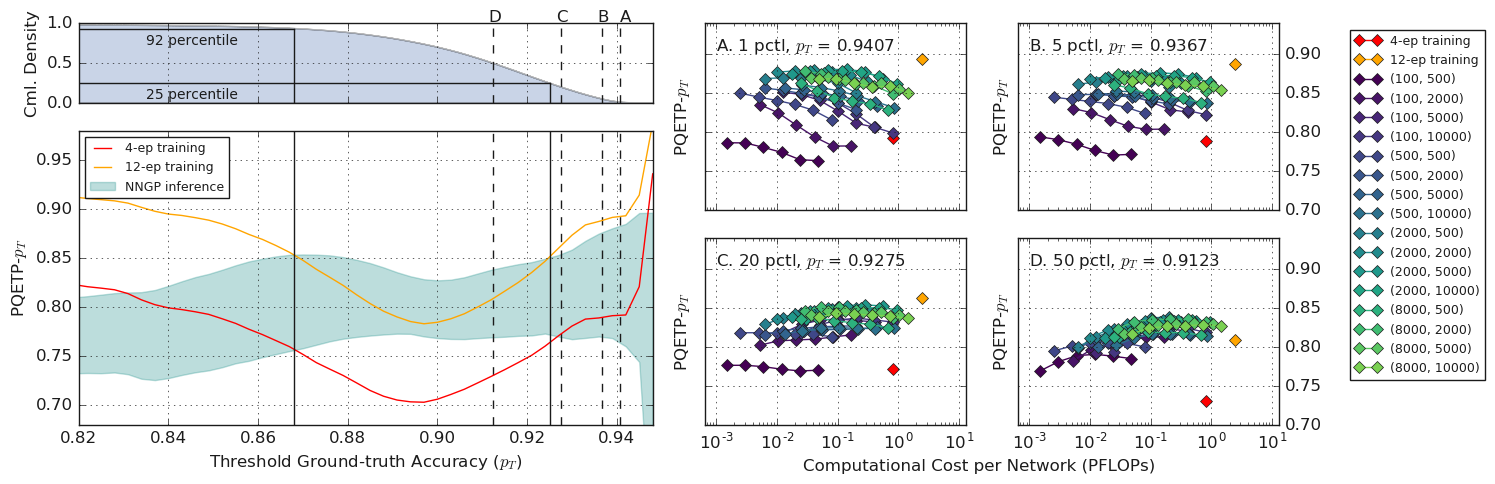}
\caption{PQETP-$p_T$ of validation accuracies from 4, 12-epoch training and NNGP inference for NAS-Bench-101. Color for NNGP inference is coded by $(N_\train, N_\val)$. ({\bf left}) PQETP-$p_T$ plotted against threshold performance $p_T$. For NNGP inference, the upper-bound and lower-bound values with respect to all configurations are shown. Cumulative density of networks above threshold $p_T$ is plotted for reference. NNGP inference stays competitive with 12-epoch training and universally better than 4-epoch training for the {25} to {92} percentile values of $p_T$ (solid lines). ({\bf right}) PQETP-$p_T$ plotted against computational cost for selected fixed values of $p_T$. These are obtained from slices of the left figure by taking the top 1, 5, 20, 50 percentile threshold values $p_T$ (dotted lines on left).}
\label{f:AUROC}
\end{figure}

In figure \ref{f:AUROC} we plot PQETP-$p_T$ for a range of threshold accuracies $p_T$ computed for validation accuracy via 4, 12-epoch training and NNGP inference. We see that while 12-epoch training validation accuracy is better suited than NNGP measures for discerning whether a network performance is within the top 1, 5 or 20 percentile, many of the NNGP measures are better at predicting whether a network performance is above-median. Meanwhile, we find that most NNGP measures have better PQETP-$p_T$ compared to 4-epoch training beyond 92 percentile values of $p_T$, despite costing less to compute.

\begin{figure}[h!]
\centering
\includegraphics[width=0.6\columnwidth]{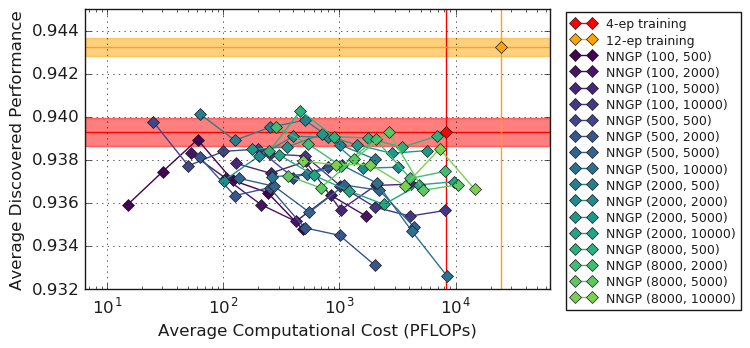}
\caption{Average discovered performance for 10 randomly sampled sets of 10k networks from NAS-Bench-101. The standard errors for 4-epoch and 12-epoch training are shown in bands.}
\label{f:nasbench top}
\end{figure}

We also compute the average discovered performance for 10 randomly sampled 10k-subsets of networks for each metric. These values have been plotted in figure \ref{f:nasbench top} against the computational cost. The standard error for the result of 4 and 12-epoch training has been plotted in bands. We see that the NNGP performance is at most as good as 4-epoch training, and is worse than 12-epoch training.

\subsubsection*{Comments on Biases of NNGP performance}

\begin{figure}[h!]
\centering
\includegraphics[width=0.95\columnwidth]{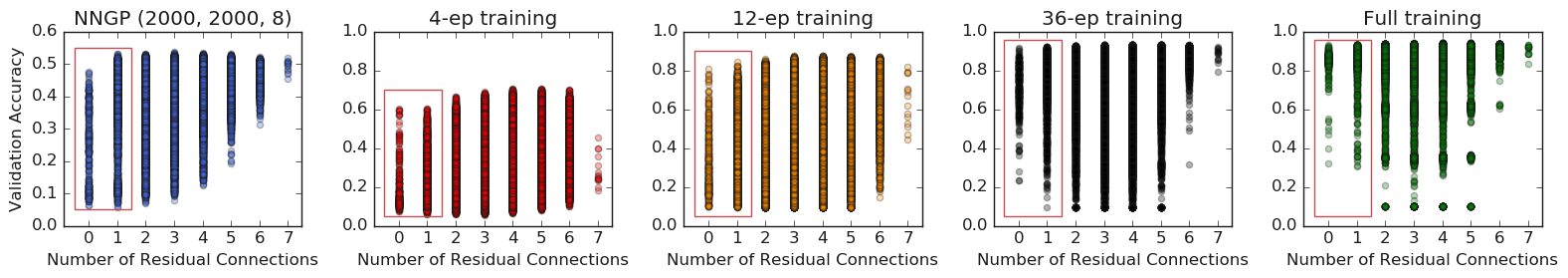}
\caption{Architectural bias of NNGPs observed on NAS-Bench-101. Relative performance of architectures with a small number of residual connections are poor early in training but improve significantly with extensive gradient-based training.}
\label{f:nngp_bias}
\end{figure}

We found that competitive architectures with poor NNGP performance were mostly ``linear" networks, having none or a small number of residual connections. In figure~\ref{f:nngp_bias} we demonstrate this bias. In the directed adjacency matrix specifying the architecture, the $k$-th diagonal elements with $k>1$ correspond to residual connections. We observe that relative performance of architectures with a small number of residual connections are poor early in training but improve significantly with extensive training. For such architectures, NNGP predicts poor performance.

\subsubsection*{Comments on usage of NTK}

NAS aims to find architectures that train well by gradient descent. Since the neural tangent kernel (NTK)~\cite{Jacot2018ntk} characterizes gradient descent training of infinitely wide networks, one might expect signals provided by NTK to be better than NNGP for NAS. To compare their utility, we computed the empirical NTK validation accuracy for a size-1k subset of networks from NAS-Bench-101 with $(N_\text{train}, N_\text{val}) = (100,1000)$. While the NTK validation accuracy has a non-trivial peak value of Kendall's Tau at $0.3150$ against the ground-truth performance for $n_\text{ensemble} = 1$, this value is lower than that computed for NNGP performance at $0.3361$ for equivalent parameters. Moreover, for the same dataset sizes, computing NTK inference is more expensive than NNGP since the Jacobian with respect to the network parameters need to be computed for all datapoints. This incurs a similar compute cost for computing gradients for all samples. In practice, computing the full Jacobian also consumes a large amount of memory, and the Vector-Jacobian Product and the Jacobian-Vector Product need to be utilized (see \texttt{nt.empirical\_ntk\_fn} in~\cite{novak2019neural} for a reference).

\begin{figure}[h!]
\centering
\begin{subfigure}{.45\columnwidth}
  \centering
  \includegraphics[width=.85\columnwidth]{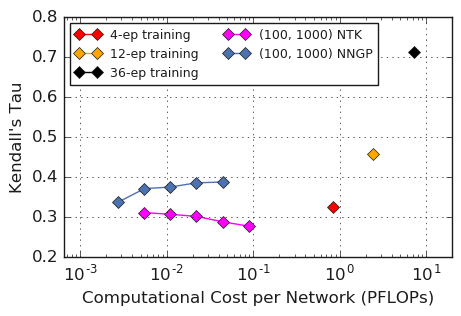}
  \label{figf:screened}
\end{subfigure}%
\begin{subfigure}{.45\columnwidth}
  \centering
  \includegraphics[width=.9\columnwidth]{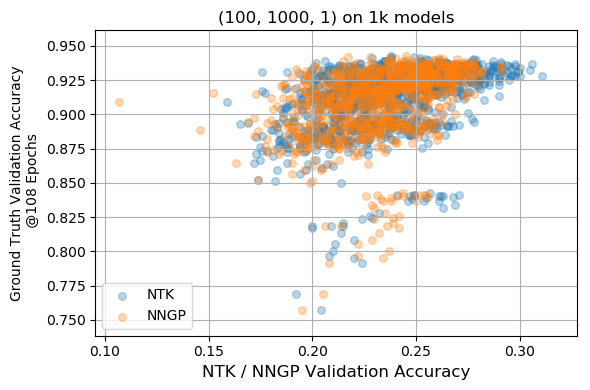}
  \label{f:hybrid}
\end{subfigure}
\caption{Comparison of NTK versus NNGP on NAS-Bench-101. NTK is not as competitive as NNGP in terms of Kendall's Tau for predicting ground truth performance. Bigger $n_\text{ensemble}$ improves the predictive quality of NNGP, while the optimal ensemble number for NTK is 1.}
\label{f:ntk}
\vskip -0.1in
\end{figure}

\subsubsection*{NNGP Performance and Model Size}

The models within NAS-Bench-101 have widely varying sizes spanning over an order of magnitude---the smallest model has $\sim$2M parameters, while the biggest one has $\sim$50M. Given this range, model size is a strong indicator of performance for NAS-Bench-101, as we explore further in section \ref{ap:size} of the SM. One may thus be concerned that NNGP performance is merely capturing the size of the network, which is a trivial aspect of the neural network architecture.%
\footnote{One can carry out an equivalent analysis of the computational budget, rather than the number of parameters of each model. For NAS-Bench-101, the model size and computational budget almost perfectly correlate---we thus restrict our discussion to model size with this understanding.}

\begin{figure}[h!]
\centering
\includegraphics[width=.9\columnwidth]{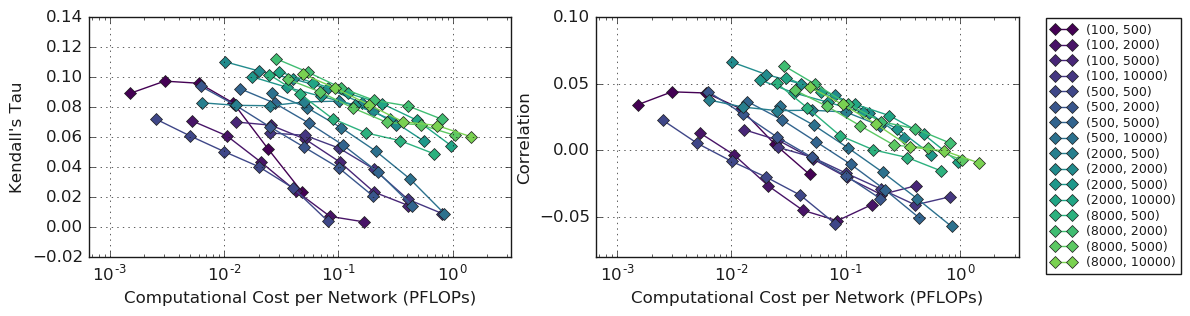}
\caption{Kendall's Tau and Pearson correlation between NNGP validation accuracy and model size plotted against the computational cost to compute the NNGP validation accuracy per network for all networks in NAS-Bench-101. NNGP validation accuracy and model size exhibit low correlation, and can thus be treated as independent predictors of model quality.}
\label{f:nngpvsize}
\vskip -0.1in
\end{figure}

In figure \ref{f:nngpvsize}, we present the Kendall's Tau and correlation between the NNGP validation accuracies and network size. We find very low correlation. We present further analysis on the model size distribution of NAS-Bench-101 and the utility of performance predictors under model size constraints in section \ref{ap:size} of the SM.

\subsection{ImageNet on MNAS Search Space}

\begin{figure}[h!]
\centering
\includegraphics[width=0.95\columnwidth]{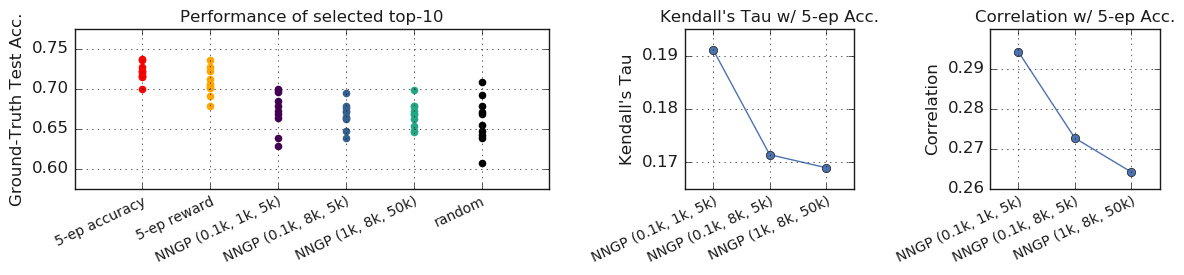}
\caption{Result of architecture selection based on 5-epoch training and NNGP inference performance from 10k samples drawn from the MNAS search space. Datasets used for NNGP inference are parameterized by $(L, N_\train, N_\val)$, where $L$ is the number of subsampled labels.}
\label{f:mnas}
\vskip -0.1in
\end{figure}

In the left-most panel of figure \ref{f:mnas}, we plot the performance of ten randomly selected networks, and the top-10 networks selected according to the 5-epoch MNAS reward, 5-epoch accuracy and NNGP validation accuracy indexed by the number of subsampled labels, training set size and validation set size, i.e., $(L, N_\train, N_\val)$. We find that the best networks according to NNGP validation accuracies perform worse than the best randomly selected network. As we discuss in section \ref{sec discuss}, this suggests that while NNGP provides a strong signal for whether a network will perform reasonably, it does not on its own identify the top performing networks.

In the latter two panels of figure \ref{f:mnas}, we give Kendall's Tau and correlation between the MNAS reward function computed after 5-epoch training and the NNGP validation accuracies. We see that there is a non-trivial correlation between the two different type of measures.

\section{Discussion and exploration of practical use of NNGP inference in NAS}
\label{sec discuss}
On NAS-Bench-101, we find that Monte-Carlo estimated NNGP inference provides a computationally inexpensive signal that shows comparable utility against the validation accuracy obtained from shortened training. We further find that NNGP inference provides a strong signal for discerning whether a network exceeds median performance, but lags behind shortened training when predicting the hierarchy of top-performers.

This is further exemplified in the experiments conducted in the MNAS search space, where a randomly sampled network already exhibits good performance. To see this point, we note that the worst 5-epoch training validation accuracy we obtain from the 10k networks we sampled from the MNAS search space is 23.69\%. This is to be compared with the NAS-Bench-101 networks, for which the average of the worst validation accuracy over 10 sets of 10k random networks after training for 4, 12, 36 and 108 epochs is given by 4.37\%, 9.12\%, 9.49\% and 9.49\%, respectively. The quality of the MNAS search space is evident, even before considering the fact that ImageNet is a much more difficult task than CIFAR-10 with a hundred times more labels. Thus the fact that the max performance over the top 10 networks ranked by NNGP is less than that over 10 random networks, despite the NNGP performance being correlated with the much more predictive shortened-training performance, is consistent with what we have observed for the NAS-Bench-101 dataset.

Based on these results, we suggest two ways that NNGP inference can be utilized in NAS. The first is that it can be used to shrink a large architecture search space in which there is a large variance in performance of networks (e.g., \cite{radosavovic2020designing}) at a low computational cost. The second is that it can be used as a complementary signal that can improve training-based performance measures.

\subsubsection*{Example of Search-Space Reduction}
Consider 10 randomly selected subsets of the networks in NAS-Bench-101 of size 10k. We compare the average discovered performance on these sets obtained by shortened training on a reduced search space obtained by selecting the top-$p$\% of networks according to NNGP validation accuracy. As baselines, we consider the average discovered performance of shortened training without NNGP screening, as well as the average discovered performance via shortened training on a random p\% subset. We experiment with $p=10$ and $30$.

\begin{figure}[h!]
\centering
\includegraphics[width=0.95\columnwidth]{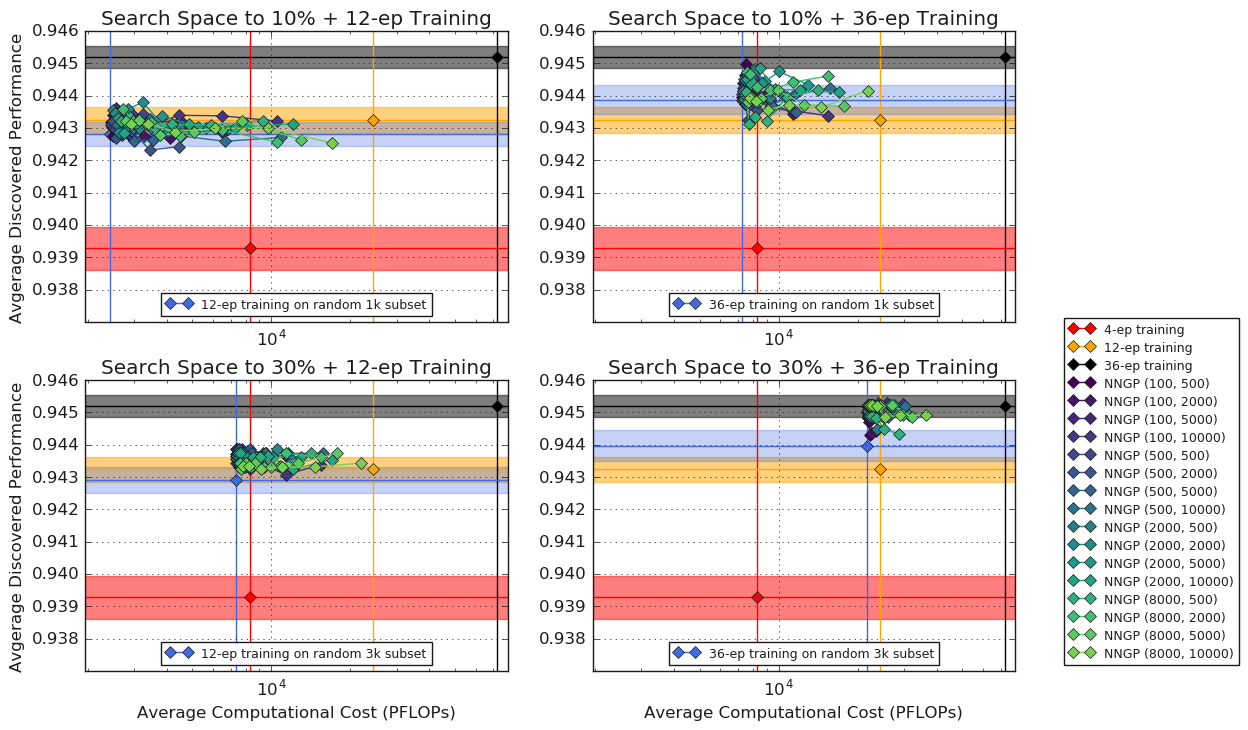}
\caption{Average discovered performance on 10 sets of 10k networks by search space reduction using NNGP validation accuracy for shortened training plotted against computational cost (viridis). The average discovered performance from 4, 12 and 36-epoch training are plotted in red, orange and black. Results from search space reduction by random selection have been plotted in blue. All networks are in NAS-Bench-101. Standard errors are shown in bands. We see that search space reduction to 30\% using NNGP performance retains discovered performance quality with lower computational cost.}
\label{f:screened}
\vskip -0.1in
\end{figure}

The results are shown in figure \ref{f:screened}. We find that 70\% reduction of the search space by screening with NNGP performance for both 12-epoch and 36-epoch training-based random architecture search does not sacrifice the performance of the search while significantly reducing the computational cost. This is to be contrasted with when the search space is reduced by random selection, which leads to marked degradation in performance (orange/black vs. blue in plots). On the other hand, 90\% reduction leads to average performance degradation, showing similar results obtained from random reduction of the search space. This is consistent with the PQETP-$p_T$ results, where we found NNGP validation accuracy to be competitive against 4 and 12-epoch training for discerning performance threshold values in the top 29 to 92 percentile. We thus expect performance degradation when the search space size is reduced to being significantly below 29\% of the original size.

\subsubsection*{Potential Hybrid Performance Metrics}
\begin{figure}[h!]
\centering
\includegraphics[width=0.7\columnwidth]{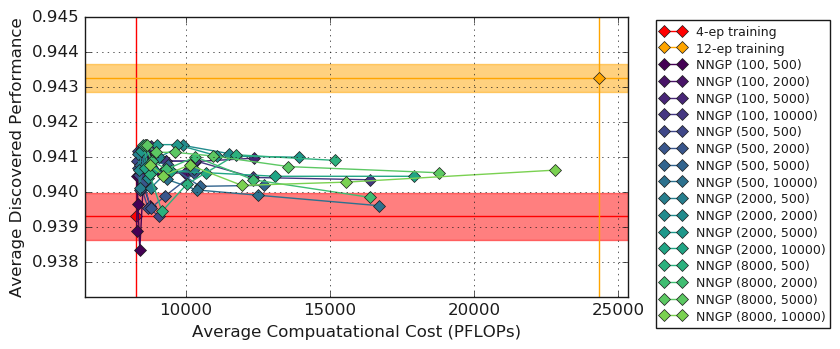}
\caption{Average discovered performance on 10 sets of 10k networks using hybrid metrics constructed from 4-epoch and NNGP performance (viridis). The average discovered performance from 4 and 12-epoch training are plotted in red and orange and black, with standard errors shown in bands.}
\label{f:hybrid}
\end{figure}

Here, we show that a simple linear model combining 4-epoch training validation accuracy and NNGP validation accuracy produces a better performance metric than 4-epoch training alone, for only a small additional computational cost. We note that we have omitted the computational cost to actually fit the linear model used, as we aim to demonstrate the existence of a hybrid performance metric with improved predictive ability.

We use a linear model with three parameters (including the bias) to fit the 12-epoch validation accuracy against the 4-epoch validation accuracy and each NNGP validation accuracy. By doing so, we obtain a hybrid performance metric, with which we measure the average discovered performance for 10 randomly selected size-10k sets of networks. The results obtained for the hybrid metric is plotted in figure \ref{f:hybrid}. We see that hybrid metrics built out of NNGPs with larger training sets can exhibit statistically significant performance gain with marginal additional computational cost.

\section*{Broader Impact}

Our research aims to reduce the computational cost of evaluating neural network performance and neural architecture search, which would lead to reduction of the environmental footprint of deep learning research and applications~\citep{strubell2019energy}.

\begin{ack}
We thank Yasaman Bahri, Gabriel Bender, Pieter-Jan Kindermans, Quoc V. Le, Esteban Real, Samuel S. Schoenholz and Mingxing Tan for useful discussions.
\end{ack}

\small

\bibliography{references}
\bibliographystyle{unsrtnat}
\normalsize
\onecolumn
\clearpage
\appendix

\begin{center}
\textbf{\large Supplementary Material}
\end{center}

\setcounter{equation}{0}
\setcounter{figure}{0}
\setcounter{table}{0}
\setcounter{page}{1}
\setcounter{section}{0}

\renewcommand{\theequation}{S\arabic{equation}}
\renewcommand{\thefigure}{S\arabic{figure}}
\renewcommand{\thetable}{S\arabic{table}}

\section{Comments on Batch-Normalization}

All convolution cells in NAS-Bench-101 and MNAS utilize batch normalization to ensure that the search space contains ResNet and Inception-like cells~\cite{ying2019nasbench101}, whose parameters need to be initialized. We initialize the moving averages in batch-normalization layers using one-forward pass on a random subset of the training set of batch size of $250$ (or $100$ for $N_{\train}=100$). In practice, we use the NAS-Bench-101's default batch-normalization momentum value of $0.997$ and use inference mode (\texttt{training=False}) to compute NNGP kernels, thus the batch normalization parameters are not far from the initial values set at zero mean and unit standard deviation.

\begin{figure}[h!]
\centering
\includegraphics[width=\columnwidth]{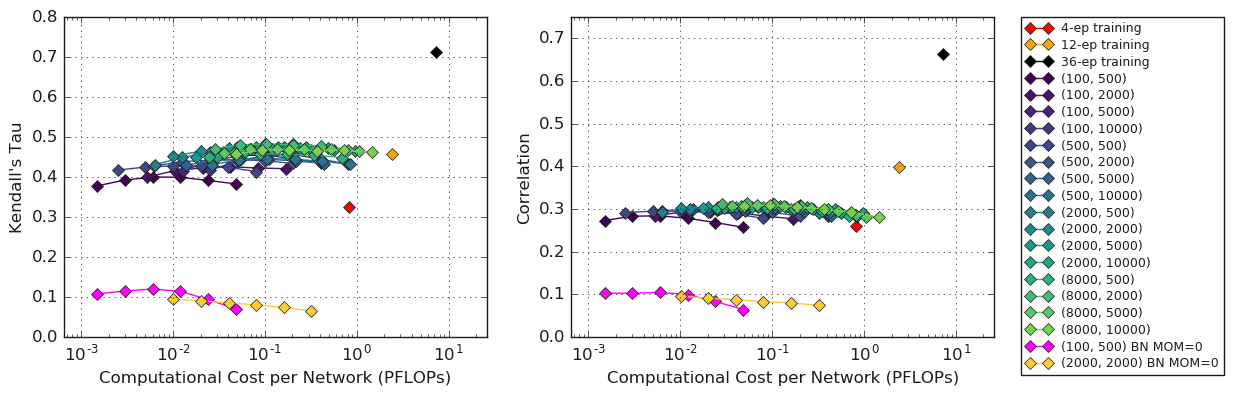}
\caption{Kendall's Tau and correlation results for NNGP with batch norm momentum set to zero.}
\label{f:bn}
\end{figure}

In figure \ref{f:bn}, we compared the result of NNGP inference with settings used in the paper, and that of updating the batch-normalization parameters with statistics of a random subset of the training set by setting momentum parameter to be $0.0$. We observe that while the performance of NNGP increases in expectation, the produced validation accuracy becomes less indicative of the ground-truth performance of the network. We in fact see that Kendall's Tau and the correlation with respect to the ground truth performance when momentum is set to zero is close to $0.1$, which is far below respective numbers for default momentum or partial training, indicating nearly random chance of predicting the correct ground truth ordering.

\section{Computational Cost}
\label{ap:cc}

The computational cost of NNGP inference stems from computing the Monte Carlo NNGP kernels $K^{tt}$, $K^{vt}$ (see algorithm 1) and from performing inference with $r$ different choices of the regularizer.

Let $F_A$ be the architecture ($A$) dependent number of flops for inference on a single sample, $d_A$ the dimension of the feature space of the penultimate layer of the network, $n_\ensemble$ the ensemble number, $L$ the number of labels, and $N_\train$ and $N_\val$ the size of the training set and validation set for the NNGP, respectively. Then the computational flops required for NNGP training is given by \begin{align}
    \textrm{Kernel Evaluation Flops} &= n_\ensemble
    ( F_A  N_{\train + \val} + 2 d_A N_\train N_{\train + \val}) \,,
\end{align}
where the first term comes from forward-propagation on the network and the second term comes from matrix computations used to construct the kernels. Meanwhile, the cost of NNGP inference via Cholesky solve for $r$ distinct values of the regularizer is given by
\begin{equation}
   \textrm{GP Inference Flops} = r \left(
   \frac{1}{3} N_{\train}^3 + 2 L N_{\train}^2 L 
   + 2L N_{\train} N_{\val} \right)
\end{equation}
Adding the two terms, we arrive at the expression
\begin{equation}\label{NNGP flops ap}
\text{NNGP Flops} = n_\text{ensemble} F_A  N_{\train + \val}
+ {1 \over 3} r N_\train^3
+ 2(dn_\text{ensemble} + 2Lr ) N_\train N_{\train + \val} \,.
\end{equation}

Denoting the FLOPs required for training a model $A$ per step per sample by $G_A$, training a model for $E$ epochs and carrying out inference costs
\begin{equation}\label{training flops ap}
  \text{Training Flops} = E G_A N^\all_\train + F_A N^\all_\val \,,
\end{equation}
where we always train/validate over the full training and validation set for gradient-based training.

\begin{figure}
\centering
\includegraphics[width=.4\columnwidth]{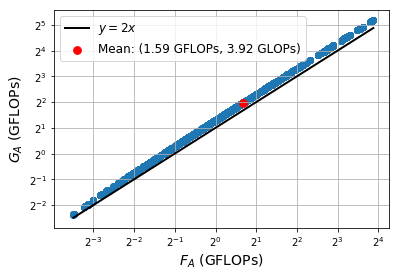}
\includegraphics[width=.4\columnwidth]{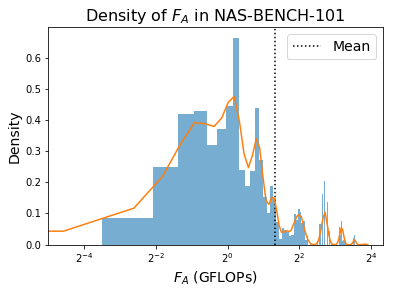}
\caption{Measured computational cost for each model in NAS-Bench-101. We observe $G_A \gtrapprox 2 F_A$ holds. }
    \label{fig:flops}
\end{figure}

We have computed $F_A$ and $G_A$ for all the networks in the NAS-Bench-101 dataset for multiple batch sizes and found that the relation $G_A \approx 2 F_A$ holds to a good approximation (see figure~\ref{fig:flops}). We thus replace $G_A$ with $2 F_A$ when evaluating computational cost.

Throughout the paper, we choose to use the computational cost per network for assessing how expensive each metric is to compute. To do so, we must evaluate $\mathbb{E}_A[F_A]$ and $\mathbb{E}_A[d_A]$ for NAS-Bench-101. We find that $\mathbb{E}_A[F_A] = 2.51$ GFLOPs (see figure~\ref{fig:flops}). Meanwhile, $d_A =512$ for all networks, which we explain in section \ref{ap:nasbench}.

\section{More Details on Search Spaces
and Experiment Design}

\subsection{NAS-Bench-101} \label{ap:nasbench}

A network in the NAS-Bench-101 dataset \cite{ying2019nasbench101} is defined by a ``cell" architecture, which is parameterized by a labelled directed graph. The labelled directed graph is defined by a set of vertices, whose label defines an operation ($3 \times 3$, $1 \times 1$ convolutions or $3 \times 3$ max pooling) and a directed adjacency matrix that specifies how to compose these operations to yield an output. The network is constructed by composing a stem convolutional layer with 128 output channels, with 3 repeated applications of blocks, which in turn consists of three concatenations of the defined cell, followed by an average pooling layer and a fully-connected layer. A down-sampling layer, which halves the height and width of the image and widens the network by twice the width is applied in between blocks.

All the networks in the NAS-Bench-101 dataset produce a feature in the penultimate layer of dimension 512. This is because each of the three blocks used for constructing the network preserve the channel number, while each of the two pooling layers situated in between the three blocks make the channel size 2 times as wide. Since the final average-pooling layer does not change the channel number, starting from the 128-channel output of the stem layer, we arrive at $128 \times 2 \times 2 = 512$ channels for the penultimate layer. This has been verified by explicit inspection of all networks.

All NNGP inference has been carried out on CPUs. All CIFAR-10 images for NNGP have been processed by standardizing the RGB channels using means 125.3, 123.0, 113.9 and standard deviations 63.0, 62.1, 66.7. No augmentations have been applied.

\subsection{MNAS Search Space} \label{ap:mnas}
The MNAS search space  \cite{tan2019mnasnet} is a factorized hierarchical search space, where the model is assumed to be composed of seven feed-forward blocks, each of which acts by repeated application of a convolutional layer. It includes multiple configuration parameters for each block, including convolution operation, kernel size, squeeze-and-excitation ratio \cite{hu2018senet}, skip operation, filter size, and repetition number (see section of 4.1 of \cite{tan2019mnasnet} for details).

The ``reward" used in MNAS search is $a \times (l_T/l)^{0.07}$ where $a$ is the validation accuracy of the model trained after 5-epochs, $l$ is the latency of the model and $l_T$ is the target latency, which is set to be $75$ milliseconds.

All NNGP inference has been carried out on CPUs. All ImageNet images for NNGP (including the NNGP training set images) have undergone standard validation processing, i.e., the RGB channels are standardized using means 123.7, 116.3, 103.5 and standard deviations 58.4, 57.1, 57.4, scaled so that the shortest edge has length 256 and then center-cropped to size 224 $\times$ 224. No augmentations have been applied.

Data processing, training and validation for 5-epoch and 400-epoch training have all been conducted in accord with \cite{tan2019mnasnet}. 5-epoch training of networks has been carried out using 8 Google Cloud TPU chips, while 400-epoch training was done using 32 Google Cloud TPU chips.

\section{NAS-Bench-101 Inference Results}
\label{ap:raw}

In this section, we present raw NNGP inference results and some basic statistics, along with the reference shortened-training results for NAS-Bench-101.

\begin{figure}[t!]
\centering
\includegraphics[width=0.98\columnwidth]{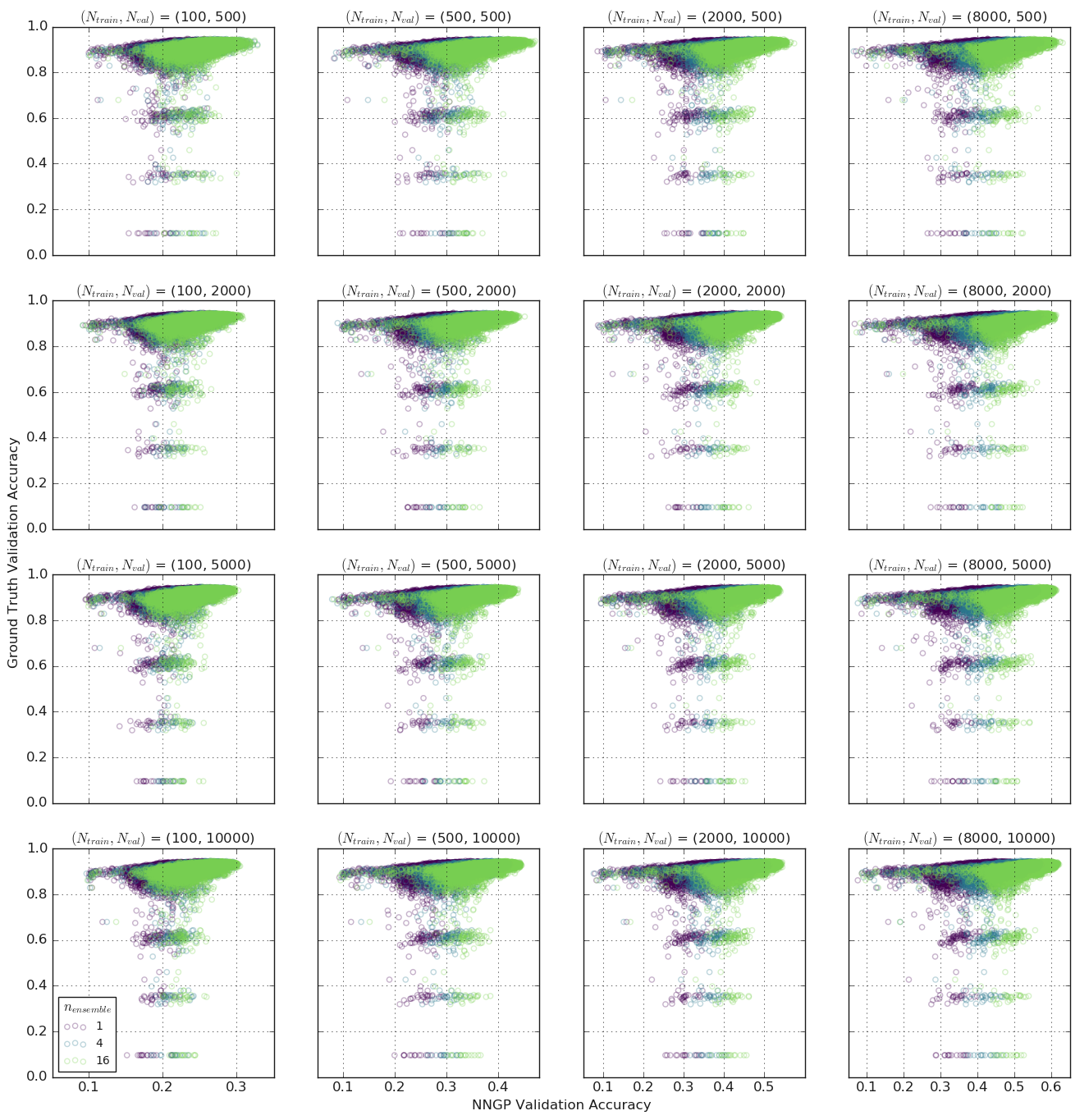}
\caption{Ground-truth accuracy plotted against NNGP validation accuracy with all values of $(N_\train, N_\val)$ for a selected set of 10k networks from the NAS-Bench-101 dataset.}
\label{f:raw all}
\end{figure}

\begin{figure}[h!]
\centering
\includegraphics[width=0.95\columnwidth]{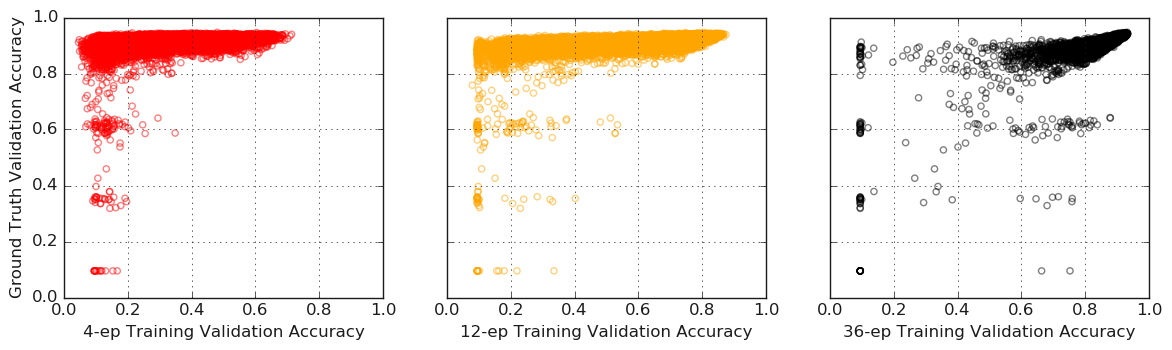}
\caption{Ground-truth accuracy plotted against shortened-training validation accuracy for a selected set of 10k networks from the NAS-Bench-101 dataset.}
\label{f:raw shortened}
\end{figure}

In figure \ref{f:raw all} we plot the ground-truth accuracy against NNGP validation accuracy for all values of $(N_\train, N_\val)$ and $n_\ensemble \in \{ 1, 4, 16\}$ for a selected set of 10k networks. The ground truth accuracy is plotted against shortened-training validation accuracies for these same networks in figure \ref{f:raw shortened}.

\begin{figure}[h!]
\centering
\includegraphics[width=0.98\columnwidth]{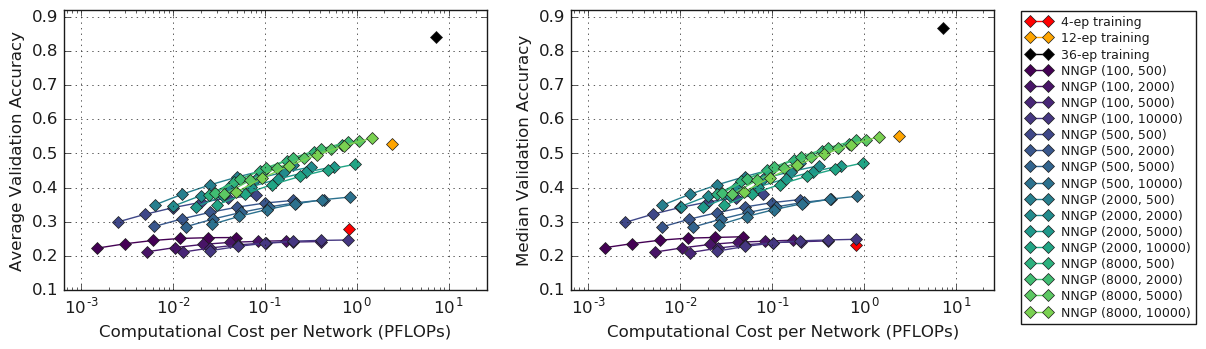}
\caption{The mean and median validation accuracy obtained from NNGP inference and shortened training plotted against computational cost.}
\label{f:basic metrics}
\end{figure}

In figure \ref{f:basic metrics} we plot the mean and median validation accuracy obtained from NNGP inference and shortened training.

\section{More Performance Measure Plots}
\label{ap:other}

In this section, we present some additional analysis on the utility of NNGP validation accuracy as an indicator of ground-truth performance.

\begin{figure}[h!]
\centering
\includegraphics[width=0.98\columnwidth]{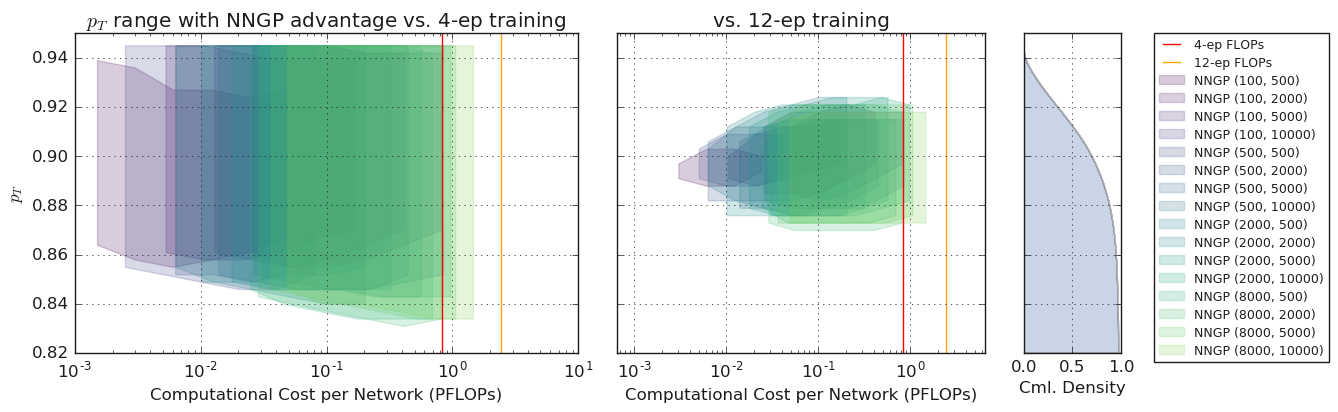}
\caption{Range of ground-truth threshold accuracies $p_T$ for which NNGP performance is a better indicator compared to 4 and 12-epoch training performance, based on PQETP. The cumulative distribution of $p_T$ is plotted for reference. Each colored shaded region indicates the area in hyperparameter space where the corresponding NNGP configuration outperforms 4-ep or 12-ep training.}
\label{f:ptrange}
\end{figure}

In figure \ref{f:ptrange}, for each NNGP validation accuracy, we plot the range of $p_T$ for which its PQETP-$p_T$ exceeds that of validation accuracies obtained from 4 and 12 epoch training. To obtain this plot, we have scanned the threshold accuracy range 0.78 to 0.95 with step size 0.003.

\section{Model Size Distribution of NAS-Bench-101}
\label{ap:size}

NAS-Bench-101 contains models with wildly varying sizes, the largest model having 20 times as many parameters as the smallest model. Given this variance, model size turns out to be a good indicator for selecting top performing models.

\begin{figure}[h!]
\centering
\includegraphics[width=0.85\columnwidth]{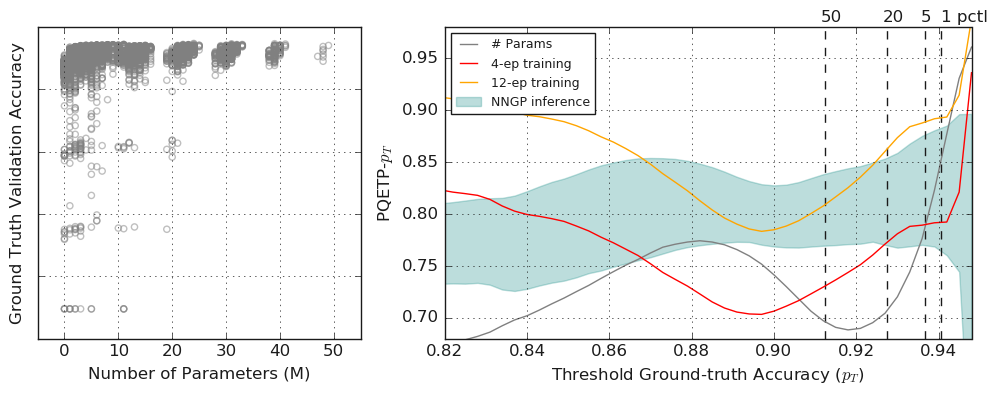}
\caption{Overview of size distribution of models in NAS-Bench-101. \textbf{(left)} Ground-truth accuracy plotted against model size for a selected set of 10k networks. \textbf{(right)} PQETP-$p_T$ of model size as a indicator for ground truth validation accuracy plotted against threshold accuracy $p_T$ in grey.}
\label{f:size overview}
\end{figure}

Figure \ref{f:size overview} presents an over-all view of the size distribution of models in NAS-Bench-101. From the left panel, which plots the ground-truth accuracy against model size for a selected set of 10k networks, we see a correlation between model size and performance. It is also evident that there are multiple clusters of models with respect to size. Meanwhile, from the PQETP plot against the threshold ground-truth performance $p_T$ depicted in the right panel, we see that model size is a very strong discriminator for threshold performances above the top percentile.

\begin{figure}[h!]
\centering
\includegraphics[width=0.98\columnwidth]{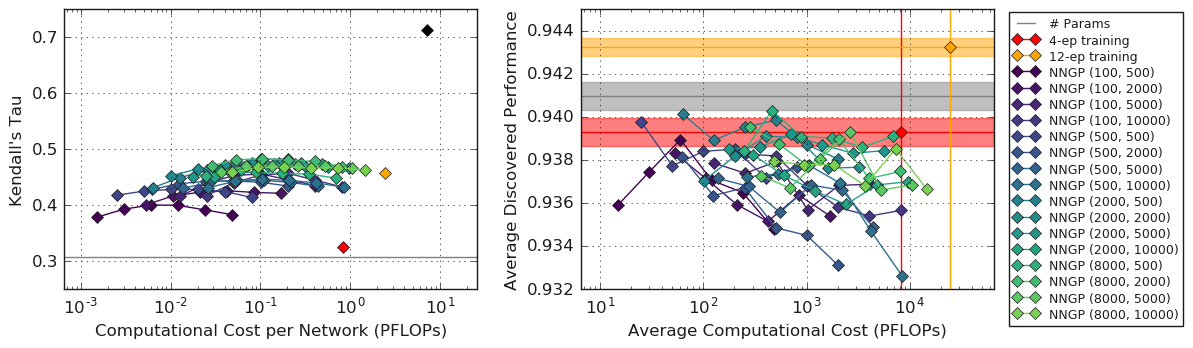}
\caption{Kendall's Tau of model size against ground truth accuracy, and average discovered performance (based on 10 random sets of size 10k) obtained by selecting biggest models plotted in grey against corresponding metrics measured for shortened training and NNGP performance.}
\label{f:size all}
\end{figure}

As a consequence, we find that while the overall ranking of performance does not align very well with model size, model size is a surprisingly good discriminator for singling out the top performing networks. This is demonstrated in figure \ref{f:size all}, which displays the Kendall's Tau of model size against ground truth accuracy, and average discovered performance obtained by choosing models based on their size.

\begin{figure}[h!]
\centering
\includegraphics[width=0.98\columnwidth]{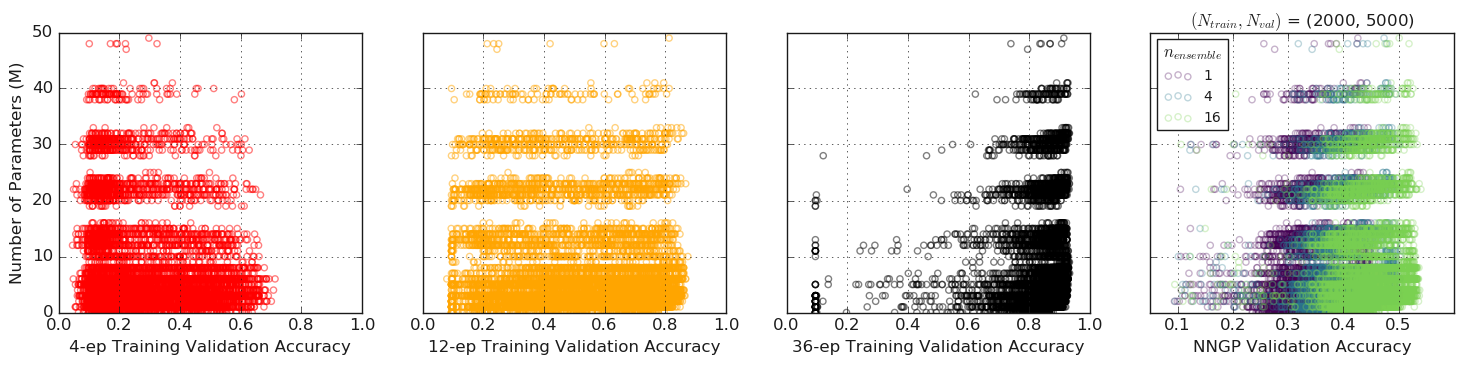}
\caption{Model size plotted against shortened-training validation accuracy as well as NNGP validation accuracy for a selected 10k subset of models. We find that there performance metrics make distinctive predictions within each model size cluster.}
\label{f:size overview 2}
\end{figure}

Given that there are distinct clusters of models with respect to their sizes that have very different number of parameters, the model size being a strong indicator of performance should be taken as a statement about the dataset, rather than a statement about the utility of model size as a performance indicator. An ideal performance metric should be able to make a distinction between models that have comparable sizes, and in fact, the goal of architecture search is often to find high-performance models with constraints on computational budget, e.g., \cite{tan2019mnasnet}. In figure \ref{f:size overview 2}, we see that indeed, shortened training performance as well as NNGP validation accuracy satisfy this criterion, having varying values within each cluster of models.

\begin{figure}[h!]
\centering
\includegraphics[width=0.98\columnwidth]{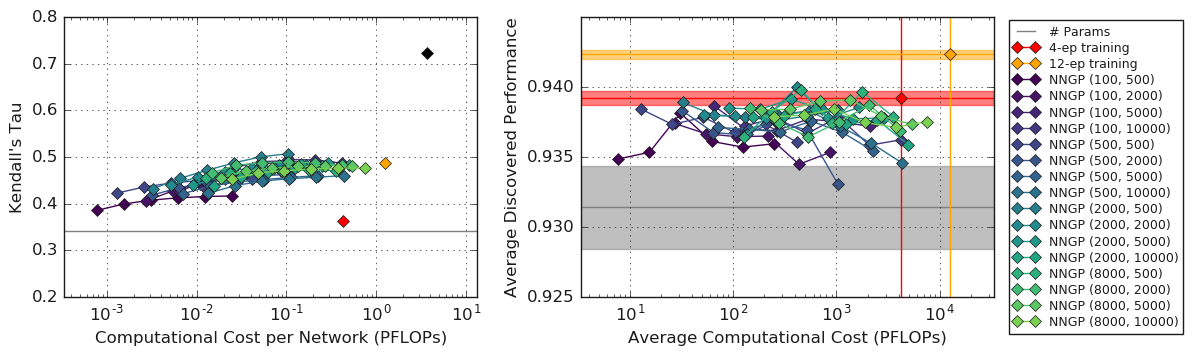}
\caption{Kendall's Tau against ground truth accuracy \textbf{(left)} and average discovered performance (based on 10 random sets of size 10k) \textbf{(right)} computed for shortened training performance, NNGP performance and model size for models with less than 10M parameters.}
\label{f:size small}
\end{figure}

To examine the utility of each performance metric within a size cluster, let us focus on the cluster of models with less than 10M parameters. There are $\sim$ 297k models in this cluster with $\mathbb{E}_A[F_A] = 1.29$ GFLOPs. As before, we compute Kendall's Tau between shortened training performance, NNGP performance and model size against ground truth performance for these models, and the average discovered performance across ten 10k-size subsets of such models according to each metric. The results are plotted in figure \ref{f:size small}. We find that the discovered performance of models selected based on model size has declined significantly more compared to that of models selected based on NNGP or shortened training accuracy in this setting. Meanwhile the hierarchy of both metrics between NNGP and shortened training performance stay largely the same.

\end{document}